\documentclass[a4paper, 10pt, conference]{ieeeconf}      

\IEEEoverridecommandlockouts                              

\usepackage{graphics} 
\usepackage{epsfig} 
\usepackage{mathptmx} 
\usepackage{times} 
\usepackage{amsmath} 
\usepackage{amssymb}  
\usepackage{multirow}
\usepackage{subfigure}

\title{\LARGE \bf
Feature Selective Small Object Detection via \\Knowledge-based Recurrent Attentive Neural Network*
}
\author{Kai Yi, Zhiqiang Jian, Shitao Chen and Nanning Zheng, Fellow, IEEE
\thanks{This research was partially supported by the National Natural Science Foundation of China (No. 61773312, 61790563), the Programme of Introducing Talents of Discipline to University (No. B13043).}
\thanks{Kai Yi, Zhiqiang Jian, Shitao Chen, Yu Chen are with Institute of Artificial Intelligence and Robotics in Xi'an Jiaotong University, Xi'an, Shannxi, P.R.China
        {\tt\small e-mail:\{yikai2015, flztiii, chenshitao\}@stu.xjtu.edu.cn}}%
\thanks{Nanning Zheng is the director of Institute of Artificial Intelligence and Robotics, Xi'an Jiaotong University, Xi'an, Shannxi, P.R.China
        {\tt\small Correspondence: nnzheng@mail.xjtu.edu.cn}}%
}  

\begin{document}

\maketitle
\thispagestyle{empty}
\pagestyle{empty}

\begin{abstract}

  At present, the performance of deep neural network in general object detection is comparable to or even surpasses that of human beings. However, due to the limitations of deep learning itself, the small proportion of feature pixels, and the occurence of blur and occlusion, the detection of small objects in complex scenes is still an open question. But we can not deny that real-time and accurate object detection is fundamental to automatic perception and subsequent perception-based decision-making and planning tasks of autonomous driving.
  Considering the characteristics of small objects in autonomous driving scene, we proposed a novel method named KB-RANN, which based on domain knowledge, intuitive experience and feature attentive selection. It can focus on particular parts of image features, and then it tries to stress the importance  of  these features and strengthenes the learning parameters of them. Our comparative experiments on KITTI and COCO datasets show that our proposed method can achieve considerable results both in speed and accuracy, and can improve the effect of small object detection through self-selection of important features and continuous enhancement of features. In addition,  we  transfer the KITTI trained model directly to the  BTSD  dataset,  and  find that KB-RANN has a good knowledge transfer ability. What’s more, we have successfully compressed and accelerated our proposed method, and deployed it in our self-developed autonomous driving car.

\end{abstract}


\section{Introduction}
    \begin{figure*}[!htbp]
            \centering
            \includegraphics[width=1.0\textwidth, trim = 0 0 0 0, clip]{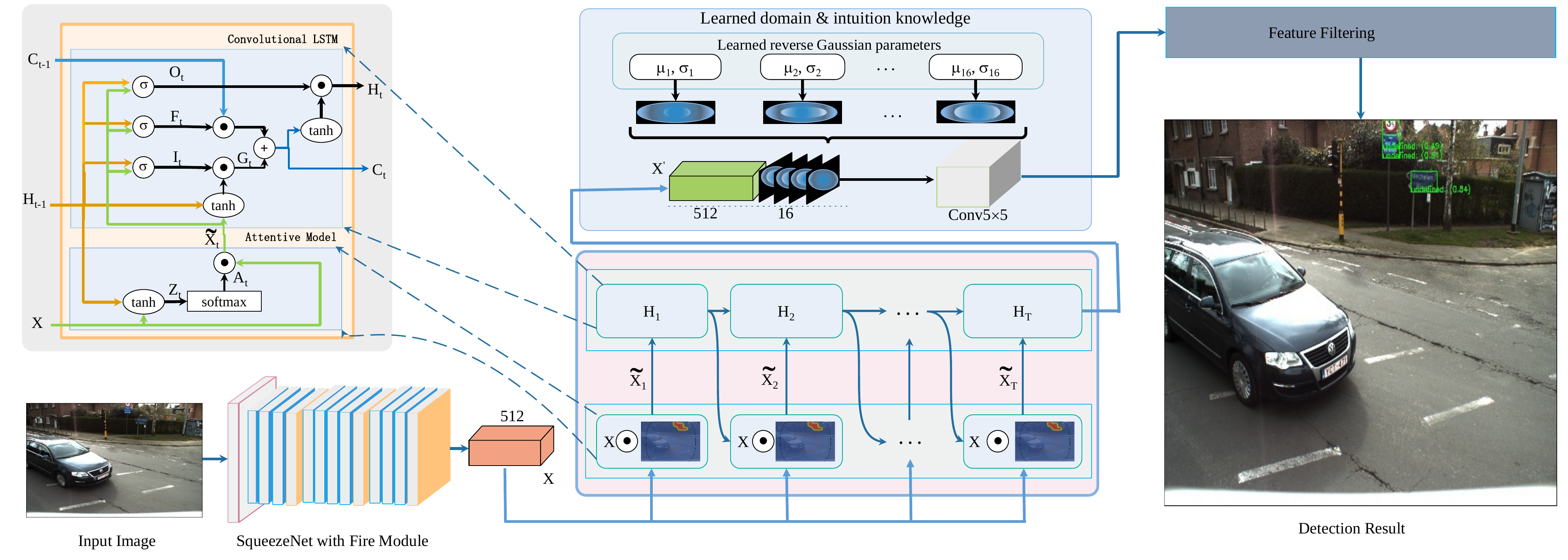}
            \caption{Flow-chart of our proposed KB-RANN. Our approach gets the feature maps of the input image, then feed it to a novel recurrent attentive neural network to improve the ability of feature representations of feature maps in a fine-grained manner. Learned domain and intuition knowledge (i.e., human gazes are the central bias of the central drivable area. The traffic signs that are always the bias of drivable this area) is injected to enhance the detection of small traffic signs. After feeding the cascaded small ConvNets with feature maps, softmax regression is used to get the final detection result.}
            \label{Fig1}
    \end{figure*}

        Vision-based object detection plays an important role in autonomous driving. At present, the general object detection methods based on deep neural network has achieved comparable or even better performance than human beings. However, due to the intrinsic limitations of deep learning and the fact that the proportion of feature pixels in the background is mall, and the target may be blurred and occluded, the detection of small objects in complex scenes is still an unresolved problem. Currently, there are two classes of object detection methods: one-stage methods and multi-stage methods. The former methods can directly obtain the coordinates and category information of the raw image in a regression manner. Nevertheless, the coarse manipulations of one-stage methods can hardly achieve considerable performance. Meanwhile, the latter methods often adopt a feature proposal network, which generate a certain number of object candidate bounding boxes, and then try to get the optimized bounding box from a large number of candidates. However, the feature proposal procession is time-consuming. In general, there is a dilemma between detection speed and accuracy. For autonomous driving scene, reliability and real-time manipulation are its basic requirements, so it is particularly important to design a detection model to meet the needs of automatic object perception of autonomous driving. This is the main focus of our paper.

        Considering that different feature layers have different ability of feature expression, low-dimensional feature maps are often used to detect smaller objects because of the difference in the size of the perceptual field of different feature maps, and high-dimensional feature maps can be used to detect larger objects. Therefore, most state-of-the-art models adopt   a multi-scale approach, and each feature layer is processed separately. Processing makes it suitable for detecting objects of different scales, so as to improve the detection accuracy of the algorithm. In addition, Focal Loss finds two reasons why the accuracy of single-stage model is difficult to improve: the proportion of positive and negative samples is unbalanced because there is no candidate box operation; the hard-negative/positive samples near GT box are difficult to train, and a new loss function Focal Loss is found. In this paper, we adopt a cognitive-inspired neural network, which can enhance the network's ability to represent features by choosing feature modules that are significant for attention and by improving the network's performance in a fine-grained manner. Experimental results show our proposed method can detect object detection well, especially small objects.

        Because the real-time requirement of the algorithm is very high for the autonomous driving sensing module, and the computing ability of the embedded system used by autonomous driving is often significantly lower than the general benchmark equipment platform for object detection, it is also important to improve the speed of the algorithm. One of the typical methods is to design more refined models, such as SqueezeNet, which can achieve AlexNet’s accuracy but at a speed of 50 times faster than the latter  by  adopting  Fire  Module.  Another is to try to reduce the model parameters and conduct channel pruning, knowledge distillation and other means to improve the speed of the algorithm. Our backbone network is designed based on SqueezeNet. Considering the uniqueness of the task and our needs, we modified the full connected layer to the full convoluted layer, resized the original kernel size, and added two extra fire modules, so that the model can detect input images of multi-scales accurately.

        For the special scene of autonomous driving, we incorporate some domain prior knowledge into the design of our detection model; the target we need to focus on is often a region outside the central position of the horizon. Details can be referred to the schematic diagram, and the application of  specific  prior knowledge will be analyzed in the structural part of the model. In addition, considering that human beings can use   the attention mechanism to select the focus areas flexibly, we propose a self-attention module, which can automatically pay attention to the possible feature region of the target, so as to detect the possible targets more efficiently. The experimental results show that the fusion of self-attention module and prior knowledge can significantly improve the detection accuracy without affecting the speed of the algorithm.

        Generally, the contribution of this paper mainly includes the following three points:

        1. In view of the particularity of autonomous driving scene, we proposed a novel model named KB-RANN based on human cognition for the first time.  Our  model  can  notice the salient features in an automatic way, and strengthen the learning of the features through iteration. At the same time, it can integrate with domain knowledge and intuitive experience. Experiments on KITTI and COCO datasets show that the proposed method can achieve considerable results both in speed and accuracy.

        2. Our proposed model has good ability to conduct knowledge transfer. The KB-RANN model parameters trained on KITTI can directly applied to the BTSD dataset. It is found that the proposed method achieved remarkable results over other algorithms.

        3. We have accelerated and optimized the proposed model and transplanted it to the embedded systems (TX2 and PX2). In addition, we successfully deployed our model in our self-developed PIONEER I autonomous driving car and tested it  in real traffic scenarios. The test results show that KB-RANN can detect traffic signs robustly at a considerable speed.

    \section{Related Work}
    \subsection{Small Object Detection}
    Due to the successful application of deep learning, great progress has been made in the field of object detection. From the practical point of view, there is still a  game  between speed and accuracy. From the point of view of detection itself, better small object detection has become the basis for further improvement of detection accuracy on multiple open source datasets. Because different feature layers of convolution networks have different perception fields, there are some related work on feature fusion of different layers: \cite{lin2017feature}, \cite{cai2016unified}. Further, \cite{zhu2019feature} proposed an anchor free strategy for feature selection, which automatically matches different bboxes to the most appropriate layer. In addition, due to the loss of original information caused by multi-pooling in high-dimensional feature layer, upsampling is also widely used \cite{fu2017dssd}.

    \subsection{Recurrent Neural Network with Attention Mechanism}
    Recursive Neural Networks (RNNs) have been widely used in sequence problem processing. From the perspective of our human brain, we pay special attention to words and visual signals that really matter to us. RNNs with this attention mechanism can achieve the same behavior, focusing on a part of the given information. An important way is to create a better designed RNNs architecture with attention mechanism. In the field of image caption, \cite{xu2015show} firstly extracts multiple features    of the input image through a convolutional neural network, and then generates the description of the image using RNNS. When it generates each word in the description, RNNs focuses on the interpretation of the relevant parts of the image. \cite{kuen2016recurrent} iteratively chooses the selected image sub-regions by using the recursive network unit, and refines them significantly in a progressive manner. \cite{Cornia17} proposes a well-designed attention LSTM architecture to refine the feature mapping extracted from convolutional networks. Based on the long-term and short-term attention neural network, we design a new recurrent attention neural network, which greatly improves the accuracy of small object detection.

    \subsection{Knowledge-based Deep Learning Systems}
    We humans can learn and perceive the world using different knowledge (for example, internal knowledge from our own experience, environmental knowledge from interactions with surrounding objects, global knowledge from the whole). Inspired by this fact, knowledge-based deep learning system has received great attention in recent years. These methods can be simply classified into two branches. One is to use intuitive prior knowledge. \cite{stewart2017label} trains  convolutional  neural  networks  to detect and track objects without any annotated samples. Another branch is the use of domain-specific knowledge. \cite{stewart2017label} proposes domain constraints to detect objects without using labeled data. In this paper, we combine domain knowledge with intuitive knowledge. We  assume that the gaze area in  the autonomous driving scene is a  drivable  area,  and  that the traffic signs always deviate from the driving area. We propose a new inverse Gauss prior distribution to formalize this problem. Compared with the non-use of the additional method, the accuracy of object detection using the fusion knowledge  is improved.

    \subsection{Knowledge Transfer}
    The knowledge learned by neural networks in a certain field can be transferred to another task. The knowledge transfer of this pattern has attracted great attention of researchers \cite{pan2010survey}, \cite{weiss2016survey}. Especially in the case that the labeled samples in the field to be migrated are not enough to train the neural network to fit, transfer learning with proper structure has been proved to be very effective. Because of the weak transfer ability of deep network, it is difficult to migrate model parameters among datasets directly to achieve satisfactory results. However, our proposed KB-RANN can automatically select salient features, which has good generalization ability for different datasets.

    \section{Model Architecture}
    In this section, we will present the complete model of our proposed KB-RANN method.

    One of our important innovations is that we proposed a novel recurrent attention neural networks, which can automatically select and fuse salient features. In addition, we can accurately extract object features at different scales through a continuous memory mechanism. Because our network is mainly oriented to the field of autonomous driving, we not only focus on the speed of object detection, but also on the accuracy of object detection. Inspired by SqueezeNet’s ability to achieve the same accuracy as AlexNet’s, but significantly faster, we modified the backbone network to make it work better for domain-specific needs.

    In addition, due to human being’s ability to use knowledge in different fields and common sense knowledge for perception, planning, decision-making and other related activities, we proposed to use some domain knowledge and intuitive knowledge in object detection. Follow-up experiments show that the relevant knowledge we introduced is very effective. What’s more, our experiments on subsequent BTSD show that our proposed KB-RANN model has good transfer ability, and the model parameters trained on KITTI can be applied to BTSD datasets with only a few necessary modifications. The details of this part will be explained in the training section.

    \subsection{Real-time Accurate SqueezeNet}
    Nowadays, many pre-trained convolutional neural network models (such as VGGNet \cite{simonyan2014very} and ResNet \cite{ResNet}) dominate the field of object detection and achieved the most advanced performance. Although these models improve the efficiency of object detection, they are time-consuming. It is very important to construct a well-designed pre-training model in real-time object detection for autonomous driving. SqueezeNet \cite{iandola2016squeezenet} is a typical pre-trained model, which has AlexNet-level accuracy but fewer parameters. In order to meet the dual needs of speed and accuracy of autonomous driving detection task, we have made some modifications to SqueezeNet model. Generally, there are two main aspects. On the one hand, the original 7 7 kernel size is replaced by two 3 3 kernel size. On the other hand, because the full convolution layer is strong enough to classify and locate objects at the same time, we introduce the fully convolutional layer in backbone. In addition, in order to improve the detection accuracy of the algorithm, we add two Fire modules at the end of  the original model, thus  setting  up a new fine-tuned backbone, which ensures the accuracy of the algorithm on the basis of guaranteeing the speed of the algorithm. The backbone network we call SqueezeNet+.

    \subsection{Recurrent Attentive Neural Network}
    Considering the speed of the model, we do not do upsampling, deconvolution and other related operations, but proceed from cognition, hoping to get a more accurate and efficient feature extraction framework. Since memory and attention mechanism is the important parts of cognitive science \cite{yi2018cognition}, we use the basic ideas of attention and memory for reference in the construction of the model, and propose a new Recurrent Attentive Neural Network, called RANN.

    Attention mechanism can help us find key feature areas from feature maps. The related structure can refer to the upper left part of 1. The attention model first obtains the input tensor X of a series of the deepest feature maps, then selects the saliency region of the input and outputs the refined outcome $\widetilde{X_t}$. The related operations are as follows:

    \begin{equation}
      \begin{aligned}
        A_t = softmax(\tanh(W_a * X + U_a * H_{t-1} + b_a) * R_a)
      \end{aligned}
    \end{equation}

    Among them, $W_a$, $U_a$, $R_a$ denotes the kernel of  the  attention model, denotes the corresponding convolutional operation, and $b_a$ is the bias.

    Next, we will analyze the memory mechanism and the design of feature refinement extraction module. In this part, we mainly refer to the long short-term memory network \cite{Schmidhuber97}, whose memory is widely used in the field of natural language processing for time-related tasks \cite{donahue2015long}, \cite{wu2016ask}. \cite{Cornia17} uses LSTM’s sequential characteristics to process features iteratively rather than using models to deal with time dependencies between inputs. We combine the characteristics of attention and memory networks and integrate them into a complete network on the left of graph 1, which we call ANN. In addition, because the high-dimensional feature map has lost the feature information of the original target through multiple pooling, we use multiple ANN modules to cascade in order to minimize the impact of feature information loss on salient feature extraction. Besides, we get output Ht from each upper ANN module and fuse it with X from the original high-dimensional feature map, and then use the result as the input of the next attention module, so that the network can automatically correct and recall the missing information from the cognitive point of view. The updating rules of the related tensors of each ANN module in our proposed RANN obeys the following equation \ref{equation2}:

    \begin{equation}
        \begin{aligned}
            &G_t = \tanh(W_g * \tilde{X_t} + U_g * H_{t-1} + b_g)\\
            &C_t = F_t \odot C_{t-1} + I_t \odot G_t\\
            &I_t = \sigma(W_i * \tilde{X_t} + U_i * H_{t-1} + b_i)\\
            &F_t = \sigma(W_f * \tilde{X_t} + U_f * H_{t-1} + b_f)\\
            &O_t = \sigma(W_o * \tilde{X_t} + U_o * H_{t-1} + b_o)\\
            &H_t = O_t \odot \tanh(C_t)
        \end{aligned}
    \end{equation}\label{equation2}

    $C_t$, $H_t$ are two typical classes of LSTM. ttt is the intermediate memory gate. In our network structure, $I_t$, $F_t$, $O_t$ are specially designed as the internal gates of attention neural networks. The combination of internal gate and intermediate memory gate realizes the correction of memory parameters. The $W_k$ and $U_k (k \in \{i, f, o, g \})$ in the model above all represent the kernel of k gates, represents the corresponding convolutional operation, and $b_k (k \in \{i, f, o, g\} )$ represents the bias o f k gates.

    The proposed RANN (cascade of multiple attention neural networks in an iterative way) can improve the detection accuracy in a gradually refined way.

    \subsection{Fusion of Domain Knowledge and Intuitive Knowledge}
\begin{figure}
        \centering\label{premodel}
        \subfigure[]{\label{a1}\includegraphics[width=0.22\textwidth, trim = 0 20 0 0, clip]{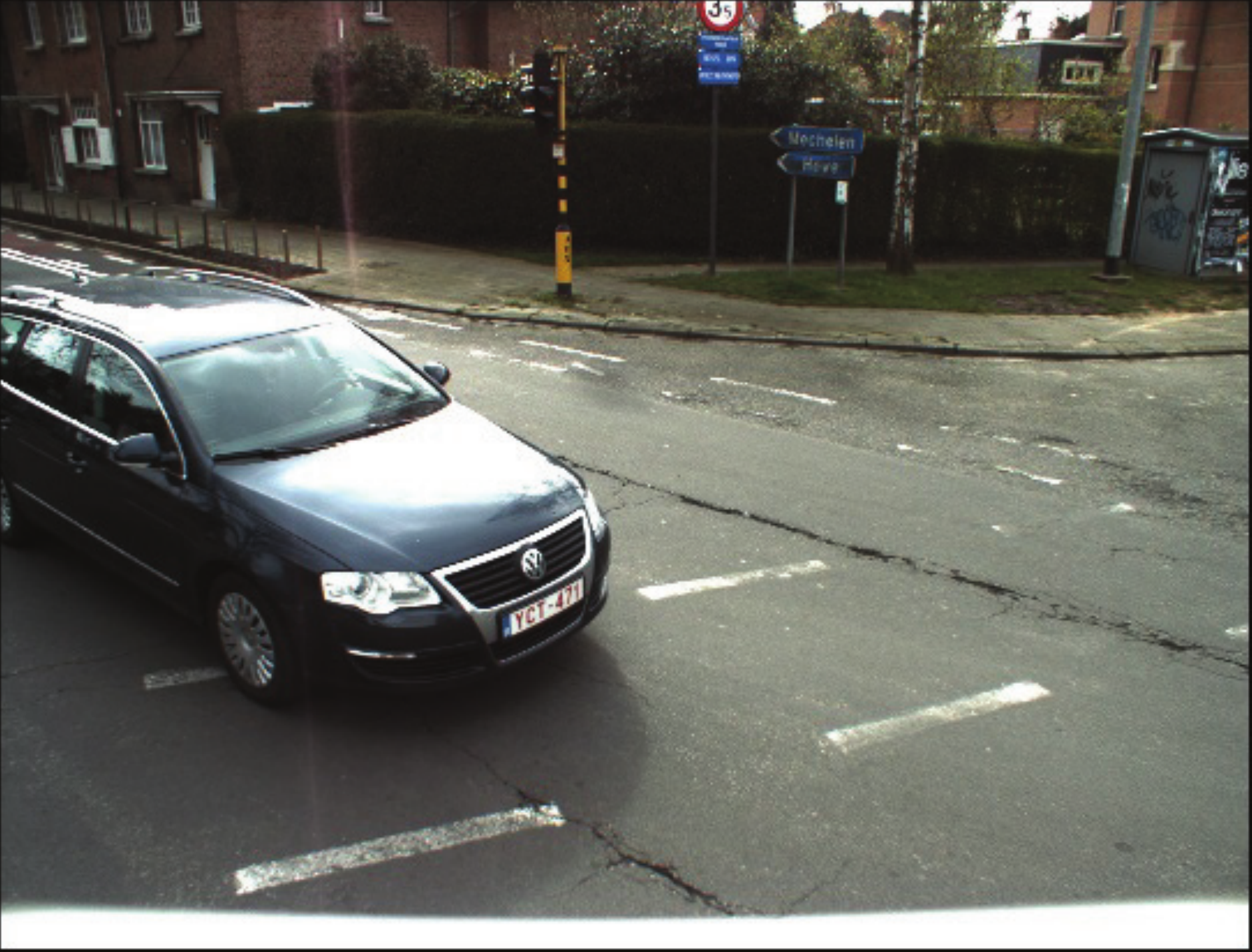}}
        \subfigure[]{\label{b1}\includegraphics[width=0.248\textwidth, trim = 3 12 3 0, clip]{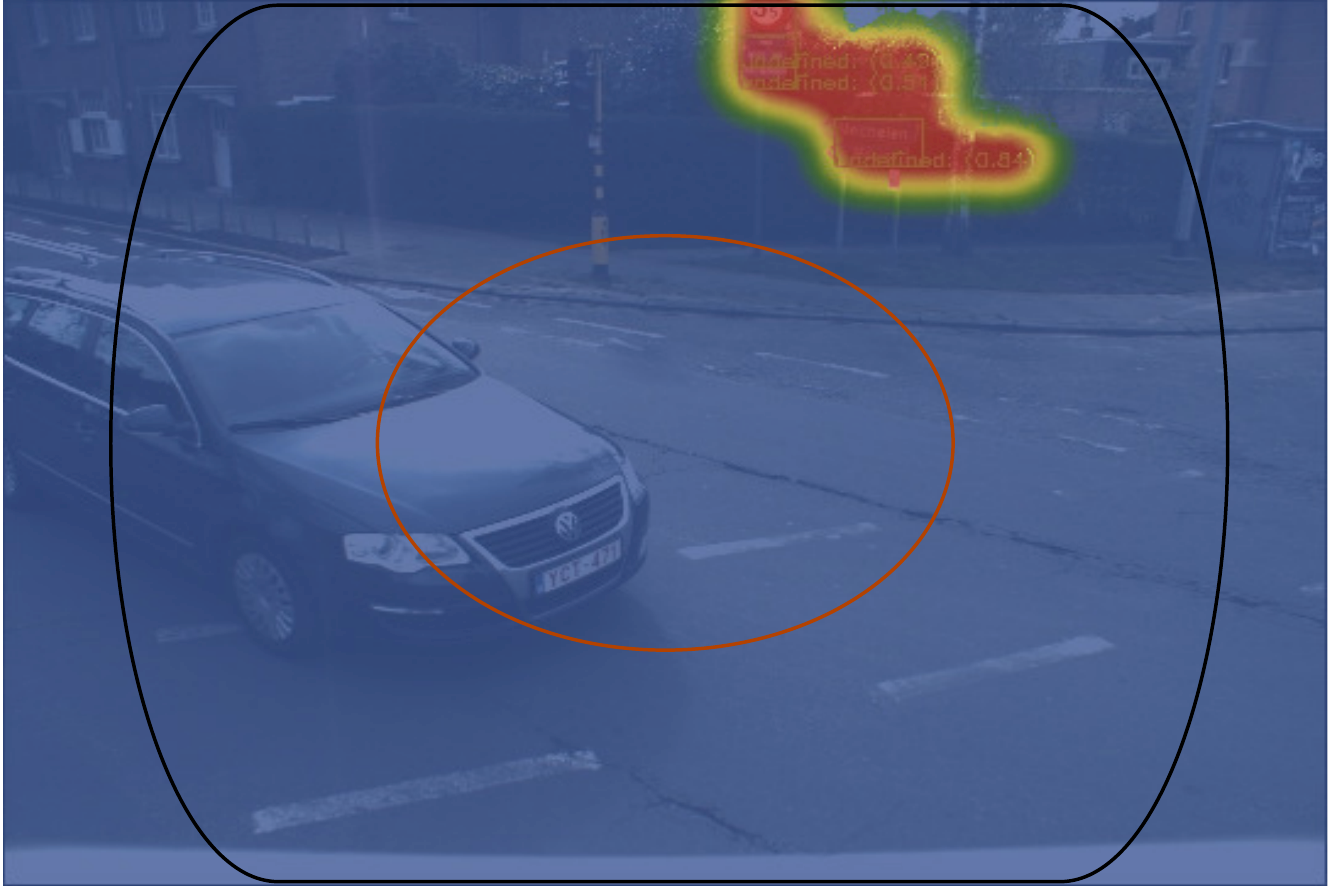}}
        \caption{(a) is the original image, (b) is the demonstration of the usage of reverse guassian prior and domain knowledge.}
    \end{figure}

    Humans can combine different kinds of knowledge in a complicated manner to solve very difficult problems. Domain knowledge is a very essential one. In the field of self-driving, people's gazes are biased toward the center. Usually, the biased center is the drivable area. In this article, we assume that the traffic signs are always located at the bias of the chosen drivable area. In Fig 2, the left figure \ref{a1} is the original image while the right Fig \ref{b1} is the demonstration of reverse guassian prior and domain knowledge. For figure \ref{b1}, the orange circle at the central location is our major attention area (i.e., the gaze of drivable area in the field of self-driving), and black circle except the drivable area is the focus of our method aimed at detecting small traffic signs. The area near the top right corner is the cluster of traffic signs.

    Prior knowledge is used to deal with the traffic signs recognition problems, but almost all of them are focused on extracting color and shape features of a traffic signs \cite{gao2006recognition}, \cite{Rinc2005Knowledge}. To the best of our knowledge, domain knowledge has not been used in dealing with traffic signs detection. The proposed reverse gaussian method is used in this task. Further, to reduce the number of parameters and facilitate the learning, we constraint that each prior should be a 2D Gaussian function, whose mean and covariance matrices are instead freely learnable. This enables the network to learn its own priors purely from data, without relying on assumptions from biological studies.

    Our proposed model can learn the parameters for each prior map through the following equation (3):

    \begin{small}
        \begin{equation}
            f^*(x,y) = \frac{1}{2\pi\sigma_x \sigma_y} \exp \left(-\left(\frac{(x-\mu_x)^2}{2\sigma_x^2} + \frac{(y-\mu_y)^2}{2\sigma_y^2}\right) \right)
        \end{equation}
    \end{small}

    And we can compute the reverse gaussian distribution by the following equation (4):

    \begin{equation}
        f(x,y) = 1 - f^*(x,y)
    \end{equation}

    We combine the N reverse guassian feature maps with those W feature maps extracted by ConvNet. Besides, we set N to 16 and W to 512. Therefore, after the concatenation, we get a mixed feature maps with 528 channels. The injection of domain and intuitive knowledge proves to be effective compared to several typical models, as we can see from the result of KB-RCNN on KITTI and COCO datasets.

    \begin{table*}[!htbp]
      \centering
      \caption{Summary of detection accuracy, inference speed of different models on KITTI object detection challenge.}\label{kitti}
      \begin{tabular}{c|c|c|c|c|c|c}
        Method & Backbone & mAP & Car & Cyclist & Pedestrain & Speed (FPS)\\
        \hline
        Faster RCNN & ResNet50 & 0.702 & 0.891 & 0.522 & 0.672 & 9.8\\
        RetinaNet & ResNet50 & 0.601 & 0.882 & 0.419 & 0.502 & 9.1\\
        SqueezeDet & SqueezeNet & 0.763 & 0.851 & 0.688 & 0.749 & 45\\
        \hline
        \textbf{RANN} & \multirow{3}*{SqueezeNet+} & 0.798 & 0.899 & 0.702 & 0.775 & 32.6\\
        \textbf{KB-RCNN} & ~ & 0.761 & 0.885 & 0.659 & 0.727 & \textbf{46.7}\\
        \textbf{KB-RANN} & ~ & \textbf{0.813} & \textbf{0.906} & \textbf{0.713} & \textbf{0.811} & 30.8\\
        \hline
      \end{tabular}
    \end{table*}

    \begin{table*}
      \centering
      \caption{Test results on MS COCO17. We used the same set of hyper-parameters to train all models in binary classification (because of the need of autonomous driving environment, we only used our models to detect pedestrains. ) The inference time includes posting processing and io operations.}\label{coco}
      \begin{tabular}{c|c|c|c|c|c|c|c|c}
        Method & Backbone & Speed (FPS) & AP & $AP_{50}$ & $AP_{75}$ & $AP_{S}$ & $AP_{M}$ & $AP_{L}$\\
        \hline
        Faster RCNN & ResNet50 & 9.9 & 0.533 & 0.830 & 0.569 & 0.355 & 0.609 & 0.702\\
        RetinaNet & ResNet50 & 9.1 & 0.518 & 0.818 & 0.541 & 0.328 & 0.597 & 0.701\\
        RetinaNet & ResNet18 & 14.5 & 0.490 & 0.798 & 0.504 & 0.309 & 0.562 & 0.685\\
        SqueezeDet & SqueezeNet & 44.5 & 0.548 & 0.842 & 0.556 & 0.359 & 0.618 & 0.741\\
        \hline
        \textbf{RANN} & \multirow{3}*{SqueezeNet+} & 39.1 & 0.557 & 0.865 & 0.612 & 0.392 & 0.659 & 0.738\\
        \textbf{KB-RCNN} & ~ & \textbf{46.3} & 0.516 & 0.762 & 0.506 & 0.315 & 0.553 & 0.669\\
        \textbf{KB-RANN} & ~ & 37.5 & \textbf{0.578} & \textbf{0.889} & \textbf{0.632} & \textbf{0.421} & \textbf{0.677} & \textbf{0.753}\\
        \hline
      \end{tabular}
    \end{table*}
    \subsection{Multi-task Loss Function}
    As \cite{redmon2016you} showed that one step training strategy that trains localization loss and classification loss together can speed up networks without losing too much accuracy, we define a multi-task loss function in the following form \ref{main-equation}:

    \begin{equation}\label{main-equation}
        \begin{aligned}
            &\frac{\lambda_{bbox}}{N_{obj}} \sum^W_{i=1}\sum^{H}_{j=1}\sum^{K}_{k=1} I_{ijk} (Q_x + Q_y + Q_w + Q_h)\\
            &+\sum^W_{i=1}\sum^{H}_{j=1}\sum^{K}_{k=1} \frac{\lambda^+_conf}{N_{obj}}I_{ijk}Q_\gamma + \frac{\lambda^-_{conf}}{WHK - N_{obj}} \tilde{I}_{ijk}\gamma^2_{ijk}\\
            &+\frac{1}{N_{obj}}\sum^W_{i=1}\sum^{H}_{j=1}\sum^{K}_{k=1}\sum^{C}_{c=1}I_{ijk}l^G_c \log(p_c).
        \end{aligned}
    \end{equation}


    The loss function contains three parts, which are bounding box regression, confidence score regression and cross-entropy loss for classification respectively. $Q*$ means $(\delta *_{ijk} - \delta *_{ijk}^{G})^2$ ($*$ represents x, y, $\gamma$, w, and h respectively).

    The first part is the loss of the bounding box regression mentioned above. ($\delta x_{ijk}, \delta y_{ijk}, \delta w_{ijk}, \delta h_{ijk}$) represents the relative coordinates of anchor-k at grid center (i,j). Meanwhile, $\delta^G_{ijk}$, or ($\delta x^G_{ijk}, \delta y^G_{ijk}, \delta w^G_{ijk}, \delta h^G_{ijk}$) is the ground truth bounding box. It can be computed by the following equation (\ref{ground-truth}):

    \begin{equation}\label{ground-truth}
        \begin{aligned}
            &\delta x^G_{ijk} = (x^G - \hat{x}_i) / \hat{w}_k,\ \
            \delta y^G_{ijk} = (y^G - \hat{y}_i) / \hat{h}_k,\\
            &\delta w^G_{ijk} = \log(w^G / \hat{w}_k),\ \
            \delta h^G_{ijk} = \log(h^G / \hat{h}_k).
        \end{aligned}
    \end{equation}

    The second part is the regression loss of confidence score. The output of the last feature map is $\gamma _{ijk}$ that represent the predicted confidence score for anchor-k corresponding to position (i, j). $\gamma ^G_{ijk}$ is the IoU of the ground truth and predicted bounding box. Besides, we penalize the confidence scores with $\tilde{I}_{ijk}\gamma ^2_{ijk}$ for those anchors are irrelevant to the task of detection. Meanwhile, $\tilde{I}_{ijk} = 1 - I_{ijk}$, $\lambda ^-_{conf}$ and $\lambda ^+_{conf}$ are used to adapt the weights.

    The last part of is the cross-entropy loss of classification. The ground truth label is $l_c^G$, and it is a binary parameter. $p_c, c \in [1,C]$ is the classification distribution that is predicted by the neural network. We used softmax regression to normalize the score so as to make sure that $p_c$ is ranged between [0,1].

    We empirically set some hyper-parameters for the above equation \ref{main-equation}. In our experiment, $\lambda_{bbox} = 5$, $\lambda^{+}_{conf} = 75, \lambda^{-}_{conf} = 100$.

    \section{Experiments}
    \subsection{Datasets and Baselines}
    We tested different models on KITTI and COCO17 datasets. KITTI is a multi-task dataset specially designed for autonomous driving scenarios. Because of its high standard IoU (0.7), it is difficult for current algorithms to achieve great performance in small object detection. Meanwhile, because there are a large number of small objects in COCO detection dataset and the occurance of blur and occlusion, the current algorithm is also relatively low in its detection accuracy. Therefore, we select these two data sets to analyze the performance of the algorithm.

    We compare the proposed KB-RANN with several popular open source algorithms in object detection, including Faster RCNN \cite{ren2015faster}, RetinaNet \cite{lin2017focal}, SqueezeDet \cite{wu2016squeezedet}. The source code we use comes directly from the original paper of the corresponding method. All we have to do is reset the RBG color mean and anchor scale for different data sets.

    \subsection{KITTI Detection Results}
    KITTI is an autonomous driving dataset. Because our requirement is object detection in autonomous driving environment, we first test our model on it. Due to the stringent requirements of model speed and detection accuracy in autonomous driving, we take both parts into consideration. At the same time, some popular classical methods are compared, including SSD, Faster RCNN, RetinaNet, SqueezeDet. In addition, we analyzed the effects of knowledge fusion and recurrent attentive neural network on the model, abbreviated as KB-RCNN and RANN respectively. The experimental results show that our algorithm achieves better results. Details can be found in Table \ref{kitti}.


    \subsection{COCO Detection Results}
    Due to the problems of too many small objects and fuzzy occlusion between image objects in MS COCOCO17 detection dataset, although some progress has been made in this data set, there is no detection algorithm with industrialized application accuracy. Because we are facing the field of autonomous driving, pedestrians are the center of our great need to focus on, so we only test the ‘person’ class. In order to verify the effectiveness of the algorithm, we not only compared Faster RCNN, RetinaNet, SqueezeDet and other popular detection algorithms, but also compared the effectiveness of different modules of KB-RANN. The specific test results on COCO  can be referred to the Table II.

    Through the above experimental results, we can know that compared with Faster RCNN, RetinaNet, SqueezeDet and other methods, our proposed KB-RANN algorithm can achieve better detection results.


    \begin{figure*}[!htbp]
        \centering
        \includegraphics[width=1.0\textwidth, trim = 0 0 0 0, clip]{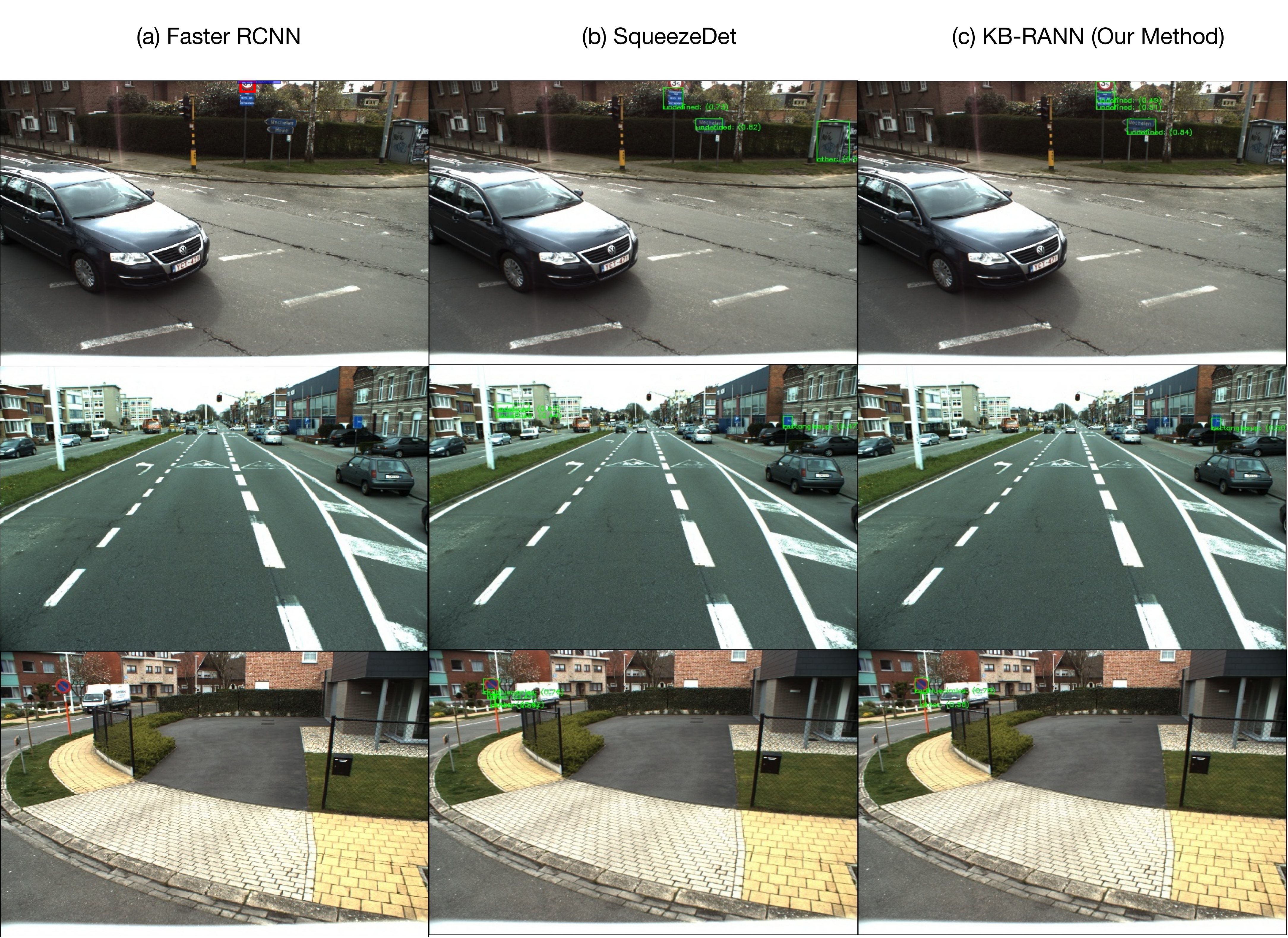}
        \caption{Comparison of detection results of three test methods, Faster RCNN \cite{ren2015faster}, SqueezeDet and KB-RANN. (a) the result of faster RCNN. The recall(miss detection) rate of this model is pretty high. As we see from the first column, this method can rarely pick out the small traffic signs from background. (b) is the result of SqueezeDet \cite{wu2016squeezedet}. During our testing, we found that the essential problem of this method is that the wrong detection is really common, it tends to select areas without traffic signs. (c) is the result of our proposed method, KB-RANN. It achieved promising detection result. Furthermore, the detecting confidence of KB-RANN is much higher than that of these aforementioned methods.}
        \label{comparison}
    \end{figure*}

    \subsection{Model Transfer}
    In order to verify the knowledge transfer ability of different models, we migrate Faster RCNN, Squeeze Det, RetinaNet and KB-RANN algorithms. We first trained the models on KITTI and then applied it directly to another dataset without additional training. Because our application needs are oriented to autonomous driving, BTSD dataset is chosen as the platform for knowledge transfer ability detection. Considering the large input scale of the original BTSD dataset, we have resized the input pictures by 1/3 by 1/3, which greatly improves the difficulty of traffic sign detection. The experimental results are shown at Table III:

    In the task of dataset transfer, because of the great difference between KITTI dataset and BTSD dataset, we use only a small sample of 1000 pictures to fine-tune the parameters of the model of KITTI dataset, so as to enhance the adaptability of the input and output process parameters of the model. The above experimental results show that the KB-RCNN network proposed by us has the fastest test speed, and the KB-RANN algorithm which integrates domain and prior knowledge and recurrent attentive module has remarkable superiority over other algorithms. In addition, since our self-developed self-driving car uses TX2/PX2 embedded platform, we have also tested the speed of all the comparative models and our proposed models. Unfortunately, all the current methods directly transplanted to TX2 can not achieve real-time results, so we subsequent use TensorRT to optimize our KB-RANN model. In addition, it is worth noting that inference time (FPS) refers to the processing time of the model itself, which does not include post processing and IO operations. Speed (TX2) contains the necessary processing operations described above.

    We visualized the output of Faster RCNN, SqueezeDet and our proposed KB-RANN at Figure \ref{comparison}. From those images, we can see that our method performs better than Faster RCNN and SqueezeDet on small traffic signs detection.

    \begin{table}
      \centering
      \caption{Results on BTSD dataset with 9 $\times$ smaller resolutions of input images (i.e. resize to 541*412).}\label{transfer-results}
      \begin{tabular}{c|c|c|c}
        Method & mAP (IoU=0.3) & mAP (IoU=0.5) & Speed (TX2)\\
        \hline
        Faster RCNN & 0.53 & 0.39 & 0.84\\
        RetinaNet  & 0.74 & 0.57 & 2.56\\
        SqueezeDet  & 0.81 & 0.61 & 9.82\\
        \hline
        \textbf{RANN} & 0.88 & 0.67 & 9.71\\
        \textbf{KB-RCNN} & 0.80 & 0.58 & \textbf{11.06}\\
        \textbf{KB-RANN} & \textbf{0.90} & \textbf{0.71} & 9.45\\
        \hline
      \end{tabular}

    \end{table}

    \section{Implementation on Embedded System}
    For a prospective and valuable autonomous driving method, there are  two  essentials  that  authentically  matter:  one  is  an available implementation on the platform of low power consumption, the other is a high frame rate to meet the requirements of real time processing. Thus, we carried out the algorithm on our self-designed embedded system, which  is aimed for evaluating and optimizing the algorithm for industrial applications such as the automobile  application.  We choose NVIDIA Jetson TX2 Module as the core of our embedded system, which balanced the power-efficient and the computing power.

    Further, we implemented the methods mentioned above (including Faster RCNN, SqueezeNet, and a series of our proposed methods), where hardware acceleration and software optimization ware carried out. The result is shown at Table \ref{transfer-results}. Our method runs on our self-designed embedded system with 10FPS, the speed is fast enough to conduct traffic sign detection. However, we are pursuing real-time performance by accelerating our method with TensorRT, which is effective in shortening the inference time of deep neural networks, as real-time detection can also be used in road-obstacle detection such as pedestrian detection.

\begin{figure}[!htbp]
        \centering
        \includegraphics[width=0.5\textwidth, trim = 0 0 0 0, clip]{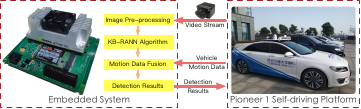}
        \caption{The implementation of the KB-RANN algorithm in our self-driving system. This figure shows the embedded system we designed based on NVIDIA TX2 to implement our algorithm on the right shows the self-driving platform we use.}
        \label{embedded}
    \end{figure}

    The proposed algorithm is implemented on the self-driving platform as shown in Fig \ref{embedded}. The Tx2 module interacts with our self-driving platform. Firstly, the system gets raw video stream from onboard camera. And then our proposed KB-RANN algorithm gets the data of detection result after image preprocessing. Next, the system fuses the motion data obtained by the self-driving platform. Finally, effective detection results can be generated and further used in decision making.

    \section{Conclusions and Future Works}
    In this paper, small target detection, especially pedestrian and traffic sign detection for autonomous driving scenes, is studied in depth. Enlightened by human cognitive mechanism, we propose a new knowledge-based recurrent attention network, which can selectively obtain salient features. At the same time, through an iterative refinement method after equalization, it can reduce the loss of features in down-sampling and wake up to complete salient features. We find that the proposed KB-RANN algorithm achieves significantly better performance than many commonly used detection methods, while guaranteeing a certain speed. In addition, from the  point of view of the application of autonomous driving, we transplant the algorithm on embedded devices, and use tools such as TensorRT to accelerate the model, and successfully deploy it on the self-developed self-driving car. At present, it can detect the traffic signs in real time.

    The future research direction may be to apply attention mechanism to traffic sign detection and pedestrian detection  in video as we can use abundant context information. More importantly, how to combine traffic rules with intuitive knowledge to construct a more real dynamic small object detection method is also the focus of our future research.

\bibliographystyle{IEEEtran}
\bibliography{bibtex}

\end{document}